\documentclass{article}
\usepackage{spconf,amsmath,graphicx}
\ninept

\usepackage{cite}
\usepackage{amsmath,amssymb,amsfonts}
\usepackage{algorithmic}
\usepackage{graphicx}
\usepackage{textcomp}
\usepackage{multirow}
\usepackage{xcolor}
\usepackage{CJK}
\usepackage{color}
\usepackage{multirow}
\usepackage{multicol}
\usepackage{makecell}
\usepackage{color}
\usepackage{xcolor}
\usepackage{colortbl}
\usepackage{bm}
\usepackage{amsmath}
\usepackage{amssymb}
\usepackage{graphicx}
\usepackage{extarrows}
\usepackage{tcolorbox}
\usepackage{CJK}
\usepackage{hhline}
\usepackage{caption}

\usepackage[colorlinks,linkcolor=black]{hyperref}
\title{MELT: Improve Composed Image Retrieval via the Modification Frequentation-Rarity Balance Network}
\name{$^{1}$Guozhi Qiu
        \ $^{1}$Zhiwei Chen
        \ $^{1}$Zixu Li
        \  $^{1}$Qinlei Huang
        \ $^{1}$Zhiheng Fu
        \ $^{2}$Xuemeng Song
        \ $^{1}$Yupeng Hu*\thanks{*Corresponding Author.}}

\address{$^{1}$ School of Software, Shandong University,\\ 
$^{2}$ Southern University of Science and Technology}

\begin{document}
\topmargin=0mm
%
\maketitle

\begin{abstract}


Composed Image Retrieval (CIR) uses a reference image and a modification text as a query to retrieve a target image satisfying the requirement of ``modifying the reference image according to the text instructions''. However, existing CIR methods face two limitations: \textbf{(1) frequency bias leading to ``Rare Sample Neglect''}, and \textbf{(2) susceptibility of similarity scores to interference from hard negative samples and noise}. To address these limitations, we confront two key challenges: \textbf{asymmetric rare semantic localization} and \textbf{robust similarity estimation under hard negative samples}. To solve these challenges, we propose the \textbf{M}odification fr\textbf{E}quentation-rarity ba\textbf{L}ance ne\textbf{T}work \textbf{(MELT)}. MELT assigns increased attention to rare modification semantics in multimodal contexts while applying diffusion-based denoising to hard negative samples with high similarity scores, enhancing multimodal fusion and matching. Extensive experiments on two CIR benchmarks validate the superior performance of MELT. Codes are available at~\href{https://github.com/luckylittlezhi/MELT}{https://github.com/luckylittlezhi/MELT}.



\end{abstract}

\vspace{-0.5em}
\section{Introduction}

\noindent 

\begin{figure}[t!]
  \begin{center}
  \includegraphics[width=\linewidth]{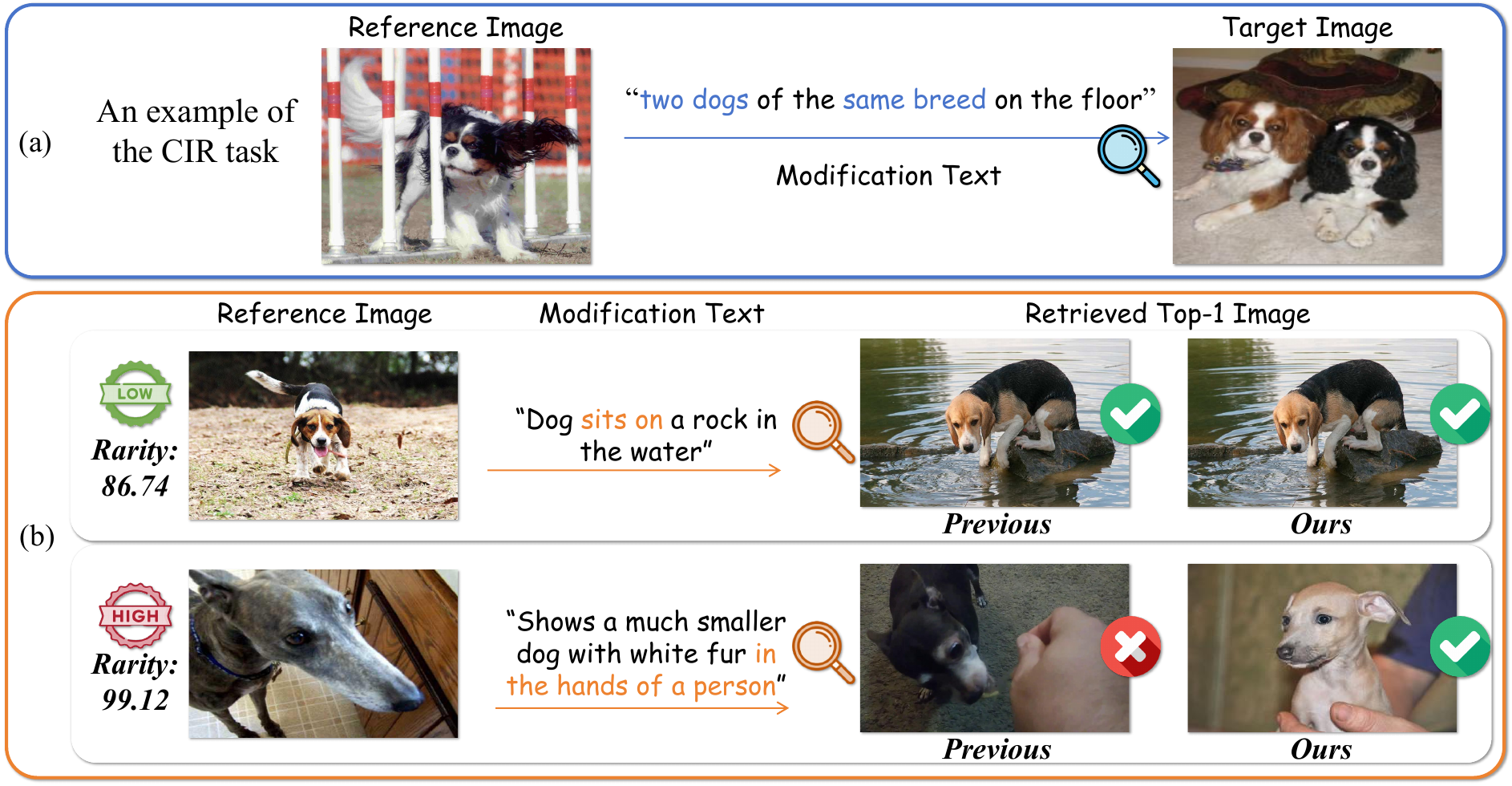}
  \end{center}
  \vspace{-20pt}
  \caption{\small An CIR example and MELT's effectiveness on rare samples.}
  \vspace{-25pt}
  \label{fig:intro}
\end{figure}

Composed Image Retrieval (CIR)~\cite{OFFSET,tgcir,INTENT,HABIT,hint} constructs a query using a reference image and a modification text to retrieve target images that satisfy the requirement of ``modifying the reference image according to the textual instructions.'' As shown in Fig ~\ref{fig:intro}(a), it is critical to emphasize that the modification text does not have strict one-to-one correspondence with the reference image in semantics. Instead, the text acts as ``modification directives'' for the reference image, requiring the model to generate a composed feature within this semantic asymmetry to accurately align target images.
Due to its interactive expression closer to real user intent, CIR holds significant application value in scenarios like video processing and visual generation~\cite{hu-tip-1,hu-tip-2,FineCIR,HUD,xu2025hdnet}, and intelligent perception~\cite{tian2025core,ReTrack,REFINE,tian2025open}.

Although significant advances in cross-modal alignment and representation learning have been achieved by existing CIR methods~\cite{tgcir,ssn}, they still suffer from two critical limitations:
\textbf{L1: Frequency bias induces rare sample neglect}. As depicted in Fig.~\ref{fig:intro}(b), the distribution of objects and modification instructions in CIR datasets is highly imbalanced. Certain objects or instructions (e.g., ``dog'', ``sitting'') exhibit high frequency, while others (e.g., ``in the hands of a person'') occur infrequently. Existing methods predominantly learn conditional probabilities dominated by high-frequency patterns, causing their retrieval results to bias toward target images containing common objects. Consequently, they neglect rare modification semantics and subtle regions during retrieval, leading to errors (e.g., retrieving images containing ``dog'' and ``hands'' while failing to satisfy the actual modification requirements).
\textbf{L2: Susceptibility to interference from hard negative samples and noise}. In specific domains like fashion, datasets often contain numerous hard negative samples with similar appearance or context. When performing single-pass scoring, the initial similarity matrix is easily distorted by amplified noise, resulting in ranking errors.

Nevertheless, addressing the above limitations remains non-trivial due to two challenges:
\textbf{(1) asymmetric rare semantic localization.}
In CIR's multimodal queries, not all words in text or regions in image correlate with modification requirements. Crucially, rare semantics affecting high-precision regions often reside in minimal image areas that easily overwhelmed by high-frequency backgrounds. The primary challenge is to precisely localize and amplify these infrequent yet critical image tokens during cross-modal interaction.
\textbf{(2) robust similarity estimation under hard negative samples.}
Estimating cross-modal similarity robustly amidst hard negative samples poses difficulties, demanding sequential denoising mechanisms that iteratively refine the initial similarity matrix toward progressive convergence, thus achieving robust composed matching.

To address these challenges, we propose a novel \textbf{M}odification fr\textbf{E}quentation-rarity ba\textbf{L}ance ne\textbf{T}work (\textbf{MELT}),which focuses more on rare modification semantics in multimodal contexts and applies diffusion denoising to high-similarity hard negative samples, enhancing multimodal fusion and matching. Specifically, MELT consists of two modules:  \textit{(a) Rarity-Aware Token Refinement} locates image regions with maximal text attention, calculates modification rarity under text guidance, and reconstructs image semantics to output composed features. \textit{(b) Diffusion-Based Similarity Denoising} employs a lightweight denoising network to perform multi-step refinement on the similarity matrix, achieving precise semantic matching.  
MELT achieved state-of-the-art results on most metrics on both FashionIQ and CIRR benchmarks. 

\vspace{-10pt}
%

\begin{figure*}[ht]
  \begin{center}
  \includegraphics[width=0.88\linewidth]{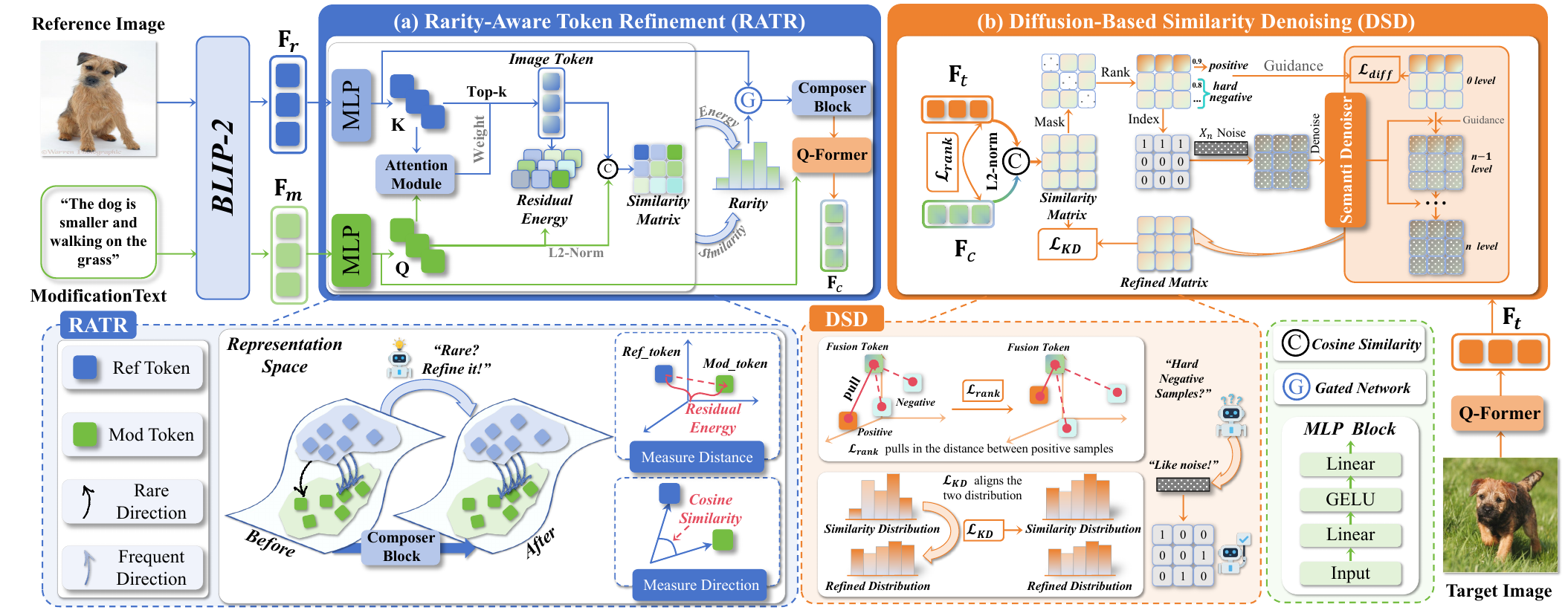}
  \end{center}
  \vspace{-20pt}
  \caption{\small Overall architecture of our proposed MELT: \textbf{(a)} Rarity-Aware Token Refinement, \textbf{(b)} Diffusion-Based Similarity Denoising.}
  \vspace{-20pt}
  \label{fig:MELT}
\end{figure*}


\section{Methodology}\label{sec:method}

\noindent

As the core innovation, our proposed MELT emphasizes rare modification semantics in multimodal contexts and employs diffusion-based denoising to handle hard negative samples with high similarity scores, facilitating multimodal composition and matching processes. Specifically, MELT consists of two modules:
\textit{(a) Rarity-Aware Token Refinement}, which leverages cross-attention to highlight key image regions indicated by text, estimates text-guided rarity, and refines local semantics based on this rarity to generate more precise composed features; and 
\textit{(b) Diffusion-Based Similarity Denoising}, which refines the initial similarity matrix via a lightweight diffusion network that progressively removes noise, yielding more robust matching against hard negatives.

\subsection{Problem Formulation}
\noindent Let $\mathcal{T}=\left\{\left(x_{r},x_{m}, x_{t}\right)_{n}\right\}_{n=1}^{N}$, $x_{r}$ is the reference image, $x_{m}$ is the modification text, and $x_{t}$ is the corresponding target image.
The objective is to construct an embedding space in which the multimodal query $(x_{r},x_{m})$ is positioned as close as possible to its associated target image $x_{t}$. Formally, the desired embedding function can be expressed as $\mathcal{F}(x_{r},x_{m}) \rightarrow \mathcal{F}(x_{t})$.

\subsection{Rarity-Aware Token Refinement (RATR)}
\noindent To measure the rarity of text-guided image modification and reconstruct image semantics based on rare parts, we design \textit{Rarity-Aware Token Refinement (RATR)} shown in Figure~\ref{fig:MELT}(a). RATR computes text attention to different image regions, combine cosine similarity to derive rarity scores, and moderately reconstruct local semantics.

Specifically, we first perform extraction and composition of multimodal representations. By Q-Former of BLIP-2~\cite{blip2}, we achieve unified extraction of reference image and modification text features,
\begin{equation}
\fontsize{9pt}{9pt}
    \textbf{F}_r=\operatorname{Q-Former}(\varPhi_\mathbb{I}(x_r)), \textbf{F}_m=\operatorname{Q-Former}(\varPhi_\mathbb{T}(x_m)),
\label{Q-Former}
\end{equation}
where $\mathbf{F}_r, \mathbf{F}_m \in \mathbb{R}^{Q\times D}$ denote the features of the reference image and the modification text, respectively. $Q$ represents the number of learnable queries and $D$ is the embedding dimension. $\varPhi_{\mathbb{I}}$ and $\varPhi_{\mathbb{T}}$ correspond to the visual and text encoder. In the same way, the target image is processed to yield its feature representation $\mathbf{F}_t \in \mathbb{R}^{Q\times D}$.

To enable semantic interaction between the reference image and modification text, we align their feature dimensions via MLP.

        \vspace{-0.5em}
\begin{equation}
    \fontsize{9pt}{9pt}
     \hat{\textbf{F}}_r = \operatorname{MLP}(\textbf{F}_r), 
     \hat{\textbf{F}}_m = \operatorname{MLP}(\textbf{F}_m), 
\label{mlp}
        \vspace{-0.5em}
\end{equation}
 where $\hat{\textbf{F}}_r, \hat{\textbf{F}}_m\in\! \mathbb{R}^{Q\times D}$.Subsequently, to facilitate subsequent fitting score calculation, we introduce a text-guided attention module to locate image regions relevant to modifications. Specifically, we treat the modification text features $\hat{\textbf{F}}_m$ as Query and the reference image features $\hat{\textbf{F}}_r$ as Key and Value, obtaining text-to-image attention weights $\textbf{w}$, formulated as:

\begin{equation}
    \fontsize{9pt}{9pt}
    \textbf{w}=\operatorname{Decoder}(Q=\hat{\textbf{F}}_m, \{K,V\}=\hat{\textbf{F}}_r),
    \label{TD}
\end{equation}
where \textit{Decoder} denotes a Transformer-decoder layer, and $\mathbf{w} \in \mathbb{R}^{Q}$ represents the correlation degree between the modification text and different regions of the reference image.

Then, for all tokens in reference image $\hat{\textbf{F}}_r$, we filter tokens based on the correlation degrees $\textbf{w}$ to obtain indices and corresponding weights of the top-$k$ image tokens most attended by modification requirements, formulated as:

\begin{equation}
\fontsize{9pt}{9pt}
\mathcal{I}=\operatorname{top}k^{{[1,Q]}} (\mathbf {w}), \quad
\tilde {\mathbf {w}}_i=\mathbf {w}_i \cdot [i\in\mathcal {I}]
\end {equation}
where $\mathcal{I}$ denotes the top-$k$ index set of reference image tokens selected based on the association weights. We then weight the corresponding image tokens with $\tilde{\mathbf{w}}$ to obtain the modification-focused image feature $\mathbf{f}_r$, formulated as:
\begin{equation}
\fontsize{9pt}{9pt}
{\mathbf{f}_r}=\sum\nolimits_{i\in\mathcal{I}}\sigma({\tilde{\textbf{w}}_{i}})\cdot\mathbf{\hat{\textbf{F}}}_{r,i},\quad i\in\mathcal{I},
\end{equation}
where $\mathbf{f}_r\in\mathbb{R}^{D}$, $\sigma$ denotes the sigmoid function, and $\mathbf{\hat{\textbf{F}}}_{r,i}$ represents the $i$-th token of the reference image feature $\hat{\mathbf{F}}_{r}$.
To evaluate the rarity of the modification, we draw inspiration from statistical methods. We introduce the residual energy based on Mahalanobis Distance, which measures the magnitude of the modification, while cosine similarity is adopted to assess its direction. The residual energy and cosine similarity are jointly used to compute a fitting score for the modification text, serving as the basis for identifying rare samples. 
Specifically, we first estimate the mean $\mu$ and covariance $\Sigma$ of the residual vector $\mathbf{r}\!=\!\hat{\mathbf{F}}_m\!-\!\mathbf{f}_{r}$ across dimensions using EMA, and then calculate the residual energy according to the covariance.

\begin{equation}
{\mu}\leftarrow\text{EMA}(\mathbf{r}),\,
{\Sigma}\leftarrow\text{EMA}\!\bigl((\mathbf{r}-{\mu})(\mathbf{r}-{\mu})^{\top}\bigr),
\label{EMA}
\end{equation}
\begin{equation}
\mathcal{E}_{\text{res}}=\sqrt{(\hat{\textbf{F}}_m-\mathbf{f}_{r})^{\top}{\Sigma}^{-1}(\hat{\textbf{F}}_m-\mathbf{f}_{r})},
\end{equation}
where $\mathcal{E}_{\text{res}}$ denotes the residual energy. We then compute the fitting score based on both the residual energy and cosine similarity,
\begin{equation}
\mathcal{R}=\mathcal{E}_{\text{res}}\cdot\bigl(1-\sigma(\cos(\hat{\mathbf{F}}_m,\mathbf{f}_{r}))\bigr),
\end{equation}
where $\mathcal{R}$ is the fitting score, and $\sigma(\cdot)$ denotes the sigmoid function, $\cos(\cdot)$ is cosine similarity. 
A larger cosine similarity indicates that the modification text is more aligned with the reference image, i.e., the induced directional change in the feature vector is insignificant.
We then compute the mean $\mu_{\mathcal{R}}$ and variance $\sigma_{\mathcal{R}}$ of the fitting score $\mathcal{R}$ using a formula similar to Eqn.~(\ref{EMA}).
For any sample, if its modification text $\hat{\mathbf{F}}^{(b)}_m$ and reference image $\hat{\mathbf{F}}^{(b)}_r$ satisfy $\mathcal{R} > \mu_{\mathcal{R}} + \gamma\sigma_{\mathcal{R}}$, we mark it as a rare sample and trigger correction, where $\gamma$ is the modification threshold. 
For such rare samples, we calibrate the top-$k$ image token set $\mathcal{I}$ most attended by the modification requirement, formulated as:
\begin{equation}
\boldsymbol{\delta}^{(b)}= \eta\,\mathbf{r}^{(b)},\;\:\hat{\textbf{F}}^{(b)}_{r,i}\leftarrow \hat{\textbf{F}}^{(b)}_{r,i}+\boldsymbol{\delta}^{(b)},\quad\forall i\in\mathcal{I}^{(b)},
\end{equation}
where $\eta$ denotes the modification intensity, which is a learnable parameter. Based on the modified reference image $\hat{\mathbf{F}}^{\prime}_r$ and the modification text $\hat{\mathbf{F}}_m$, we obtain the composed feature $\mathbf{F}_c\in \mathbb{R}^{Q\times D}$, formulated as:
\begin{equation}
\textbf{F}_c=\operatorname{Q-Former}((\hat{\textbf{F}}^{\prime}_r),(\hat{\textbf{F}}_m)).
\end{equation}

\subsection{Diffusion-Based Similarity Denoising(DSD)}
\noindent

To achieve robust similarity estimation under hard negative samples, we design \textit{Diffusion-Based Similarity Denoising (DSD)}. This module treats the influence of samples that frequently appear during training and exhibit high similarity to true labels (i.e., hard negative samples) as noise on the similarity matrix. Consequently, it employs a trained denoiser to eliminate their semantic impact.

Differing from generative tasks, we define the precise similarity matrix between composed features $\textbf{F}_c$ and target features $\textbf{F}_t$ as the learning target for noise modeling. The denoiser is trained to recover the true label distribution by removing noise introduced by hard negative samples, thereby optimizing accurate interpretation of multimodal semantics.

Concretely, we propose a semantic denoiser shown in Figure~\ref{fig:MELT}(b). After obtaining the original similarity vectors, we treat them as ``noisy signals'' because hard negative samples exhibit similarity scores close to positive samples, forming false peaks. MELT's semantic denoiser performs denoising through multi-step DDIM~\cite{ddim}. Specifically, for the $j$-th query in a batch, we place the positive sample first and sort remaining samples by similarity in descending order. The reordered indices $\mathcal{I}^{(j)}_s$ are shown below:

\begin{equation}
\mathcal{I}^{(j)}_s=\bigl[j;\;\mathop{\text{descend}}\limits_{i\neq b}\,\mathbf{S}_{i}^{(j)}\bigr]\in\mathbb{R}^{B},
\end{equation} 
where $\mathbf{S}_{i}^{(j)}$ denotes the similarity vector, and $B$ represents the batch size. Then, we rearrange both the similarity and target features as,

\begin{equation}
\tilde{\mathbf{s}}^{(j)}=\mathbf{S}^{(j)}[\mathcal{I}^{(j)}_s],\;\;
\mathbf{T}^{(j)}=\mathbf{F}^{(j)}_t[\mathcal{I}^{(j)}_s],
\label{rerank}
\end{equation}
where each slice $\tilde{\mathbf{s}}^{(j)}\in\mathbb{R}^{B}$ represents the reordered similarity vector for the $j$-th sample, and $\mathbf{T}^{(j)}\in \mathbb{R}^{Q\times D}$ denotes all reordered target features for the -th sample.

Based on similarity indices $\mathcal{I}_s$ , we obtain corresponding ground-truth labels $\mathbf{x}_0$. We construct a forward diffusion process that makes it approach the similarity matrix form while preserving its label distribution. Adopting standard Gaussian noise, the noisy matrix $\mathbf{X}_t$ at step $t$ is given by:

\begin{equation}
\mathbf{x}_t = \sqrt{\bar{\alpha}_t}\,\mathbf{x}_0 + \sqrt{1-\bar{\alpha}_t}\,\boldsymbol{\varepsilon},\quad \boldsymbol{\varepsilon}\sim\mathcal{N}(\mathbf{0},\mathbf{I}),
\end{equation}
where $\bar{\alpha}_t=\prod_{i=1}^{t}(1-\beta_i)$ is the cumulative product coefficient, and $\beta_i\in(0,1)$ denotes the noise scheduling coefficient.

We utilize the composed feature $\mathbf{F}_c$ , the similarity vector $\tilde{\mathbf{s}}$ obtained from Eqn.(\ref{rerank}), and the target feature set $\mathbf{T}$ as guiding knowledge for the denoising process. The single-step DDIM~\cite{ddim} denoising is formulated as follows:

\begin{equation}
\mathbf{x}_{t-1}=\operatorname{DDIM}\!\Bigl(\mathbf{x}_t,t;\;\tilde{\mathbf{s}},\textbf{F}_{c},\textbf{T}\Bigr).
\end{equation}
where DDIM denotes the denoising diffusion implicit model.

Unlike generative diffusion models that predict noise, our trained semantic denoiser gradually removes noise from $x_n$ to $x_0$ to obtain a purified similarity matrix. This process is constrained by the loss $\mathcal{L}_{diff}$ :

\begin{equation}
\mathcal{L}_{diff} = \frac{1}{B}\sum_{i=1}^{B}D_{\text{KL}}(\tilde{S}\,\|\,X_0),
\label{loss_diff}
\end{equation}
where $\tilde{S}$ is the reordered similarity matrix, $X_0$ is the corresponding ground-truth matrix, and $B$ denotes the batch size.

We input the target similarity matrix $S$ into the trained denoiser to eliminate noise from hard negative samples, obtaining the adjusted similarity matrix $\hat{S}$ . To promote synergistic optimization between the denoising process and model training, we introduce a knowledge distillation loss, formulated as follows:

\begin{equation}
\mathcal{L}_{KD} = \frac{1}{B}\sum_{i=1}^{B}D_{\text{KL}}(S\,\|\,\hat{S}).
\label{loss_KD}
\end{equation}

Subsequently, we adopt a widely used batch-based classification loss function to pull composed features closer to corresponding target features, formulated as follows:

\begin{equation}
\fontsize{9pt}{9pt}
\mathcal{L}_{bbc} = \frac{1}{B} \sum_{i=1}^{B} -\log \left\{ \frac{\exp \left\{ \operatorname{s} \left( \mathbf{F}_{ci} , \mathbf{F}_{ti} \right)  / \tau\right\}}{ \sum_{j=1}^{B} \exp \left\{ \operatorname{s} \left( \mathbf{F}_{ci}, \mathbf{F}_{tj} \right) / \tau \right\}  } \right\},
\label{bbc}
\end{equation}

where $B$ denotes the batch size, and ${\mathbf{F}_{ci}, \mathbf{F}_{ti}}$ represent the composed feature and target feature for the $i$-th triplet sample, respectively. Finally, we obtain the overall loss function as follows:

\begin{equation}
    \fontsize{9pt}{9pt}
    \mathbf{\Theta^{*}}=
    \underset{\mathbf{\Theta}}{\arg \min } \left( {\mathcal{L}}_{bbc}+\kappa {\mathcal{L}}_{KD}+\lambda{\mathcal{L}}_{diff}\right),
    \label{optimization}
\end{equation}
where $\mathbf{\Theta^{*}}$ denotes the learnable parameters in MELT, and $\kappa,\lambda$ are hyperparameters.

\section{Experiments}\label{sec:experiments}
\subsection{Experimental settings}\label{sec:experiment setting}
\noindent\textbf{Datasets}. We evaluate our proposed MELT on two datasets, including an open-domain dataset, i.e., CIRR~\cite{cirr} and a fashion-domain datasets, i.e., FashionIQ~\cite{FashionIQ}.

\begin{table*}[ht]
  \caption{Performance comparison on FashionIQ and CIRR relative to R@$k$(\%). The overall best results are in bold, while the best results over baselines are underlined. The Avg metric in CIRR denotes (R@$5$ + R$_{subset}$@$1$) / 2.}
  \centering
  \vspace{-10pt}
        \resizebox{\linewidth}{!}{
        
    \begin{tabular}{l|cc|cc|cc|cc|cccc|ccc|c}
    \hline
    \hline
    \multicolumn{1}{c|}{\multirow{3}{*}{Method}} &  \multicolumn{8}{c|}{FashionIQ}                              & \multicolumn{8}{c}{CIRR} \\
\cline{2-17}       & \multicolumn{2}{c|}{Dresses} & \multicolumn{2}{c|}{Shirts} & \multicolumn{2}{c|}{Tops\&Tees} & \multicolumn{2}{c|}{Avg} & \multicolumn{4}{c|}{R@$k$} & \multicolumn{3}{c|}{R$_{subset}$@$k$} & \multirow{2}{*}{Avg} \\
\cline{2-16}            & R@$10$  & R@$50$  & R@$10$  & R@$50$  & R@$10$  & R@$50$  & R@$10$ & R@$50$  & $k$=$1$   & $k$=$5$   & $k$=$10$ & $k$=$50$ & $k$=$1$   & $k$=$2$  & $k$=$3$ &  \\
    \hline
    \hline
    TG-CIR~\cite{tgcir}\scriptsize{\textcolor{gray}{(ACM MM'23)}} & 45.22  & 69.66  & 52.60  & 72.52  & 56.14  & 77.10  & 51.32  & 73.09 & 45.25  & 78.29  & 87.16  & 97.30  & 72.84  & 89.25  & 95.13  & 75.57  \\
    COVR-BLIP~\cite{covr}\scriptsize{\textcolor{gray}{(AAAI'24)}}& 44.55  & 69.03  &  48.43  & 67.42  & 52.60  & 74.31 & 48.53  & 70.25  & 49.69  & 78.60  & 86.77  & 94.31  & 75.01  & 88.12  & 93.16  & 76.81  \\

    SSN~\cite{ssn}\scriptsize{\textcolor{gray}{(AAAI'24)}} & 34.36  & 60.78  & 38.13  & 61.83  & 44.26  & 69.05  & 38.92  & 63.89  & 43.91  & 77.25  & 86.48  & 97.45 & 71.76  & 88.63  & 95.54  & 74.51  \\
    SPRC~\cite{sprc}\scriptsize{\textcolor{gray}{(ICLR'24)}}& 49.18  & 72.43  & \underline{55.64}  & 73.89  & 59.35  & 78.58  & 54.72  & 74.97  & 51.96  & 82.12  & 89.74  & 97.69  & \textbf{80.65}  & \underline{92.31}  & \underline{96.60}  & \underline{81.39}  \\
     IUDC~\cite{iudc}\textcolor{gray}{\scriptsize{(TOIS'25)}} &35.22& 61.90 &41.86& 63.52& 42.19& 69.23& 39.76& 64.88 & -     & -     & -     & -     & -     & -     & -     & -\\
    COPE~\cite{cope}\scriptsize{\textcolor{gray}{(ACL'25)}}& 39.85  & 66.98  & 45.03  & 66.81  & 48.61  & 72.01  & 44.50  & 68.60  & 49.18  & 80.65  & 89.86  & 98.05  & 72.34  & 88.65  & 95.30  & 76.49  \\
    QuRe~\cite{QuRe}\scriptsize{\textcolor{gray}{(ICML'25)}}& 46.80  & 69.81  & 53.53  & 72.87  & 57.47  & 77.77  & 52.60  & 73.48  & \underline{52.22}  & \underline{82.53}  & \underline{90.31}  & \underline{98.17}  &  78.51  & 91.28  & 96.48  & 80.52  \\
    MMF~\cite{mmf}\scriptsize{\textcolor{gray}{(ICASSP'24)}}& 40.45  & 65.29  & 44.85  & 66.29  & 49.26  & 70.98  & 44.85  & 67.52  &  \underline{24.94}  &  \underline{61.08}  & 83.78  &  \underline{56.57}  \\

    MEDIAN~\cite{median}\scriptsize{\textcolor{gray}{(ICASSP'25)}} & 46.90  & 70.30  & 52.65  & 73.96  & 57.62  & 78.63  & 52.39  & 74.30 & 45.66 & 78.72 & 87.88 & 97.89 & 75.52 & 89.45 & 95.57 & 77.12\\
    PAIR~\cite{pair}\scriptsize{\textcolor{gray}{(ICASSP'25)}} & 46.78  & 70.93  & 52.60  & 73.80  & 58.91  & 78.81  & 52.76  & 74.51 & 46.36 & 78.43 & 87.86 & 97.90 & 74.63 & 89.64 & 95.61 & 76.53\\
    ENCODER ~\cite{encoder}\scriptsize{\textcolor{gray}{(AAAI'25)}}& \underline{51.51}  & \underline{76.95}  & 54.86  & \underline{74.93} & \underline{62.01}  & \underline{80.88}  & \underline{56.13}  & \underline{77.59} & 46.10  & 77.98  & 87.16  &  97.64 & 76.92  & 90.41  & 95.95  & 77.45      \\

    \hline
    \hline
    \textbf{MELT~(Ours)}  & \textbf{53.13} & \textbf{77.54} & \textbf{59.81} & \textbf{78.99} & \textbf{64.67} & \textbf{83.06} & \textbf{59.20} & \textbf{79.86} & \textbf{52.76} & \textbf{83.04} & \textbf{90.33} & \textbf{97.94} & \underline{80.63} & \textbf{92.68} & \textbf{96.77} & \textbf{81.84} \\
    \hline
    \hline
    \end{tabular}%
    }
        \vspace{-15pt}

  \label{tab:main}%
\end{table*}%

\noindent\textbf{Evaluation metrics.}  
Following previous work, on CIRR we report Recall@$k$ ($k\!=\!1,5,10,50$), Recall$_{\text{subset}}$@$k$ ($k\!=\!1,2,3$), and the average $(\text{R}@5 + \text{R}_{\text{subset}}@1)/2$.  

\noindent\textbf{Implementation Details}. 
We use the pre-trained BLIP2~\cite{blip2} to extract features, and set the number of learnable queries $Q$ in the Q-former to $32$. We train the MELT model using the AdamW optimizer, with a learning rate of $1e-5$, hidden dimension $D$ set to $256$, and a batch size $B$ of $128$. The attention encoder used in Equation Eqn.($\ref{TD}$) is configured with $1$ layer and $12$ heads. Through grid search, we set the trade-off parameter $\kappa$ to $0.3$, $\lambda$ to $0.1$, modification strength $\eta$ to $0.8$, and modification threshold $\gamma$ to $2.2$. The model is trained on a single NVIDIA V100 GPU with $32$GB of memory.

\subsection{Perfermance Comparison}
\noindent 
We report the comparison results of the proposed MELT method in Table~\ref{tab:main}. Based on these results, we make the following two observations:
1) MELT achieves outstanding performance on both datasets. Notably, compared to the suboptimal results, MELT improves the average R@$10$ by $3.07$\% and the average R@$50$ by $2.27$\% on the FashionIQ dataset. Meanwhile, MELT improves the R@$1$ metric on CIRR, with R@$5$, and R@$10$ all showing improvements of over $0.5$\%, and the average performance increasing by $0.45$\%. This demonstrates that the method exhibits strong generalization ability in both the fashion domain and open-domain settings.
2) It is observed that MELT achieves a significantly greater performance improvement on the FashionIQ dataset compared to the CIRR dataset. This is as expected, as the fashion domain dataset typically contains rare samples from less common categories, as well as numerous difficult negative samples from popular categories.For open-domain datasets, samples generally exhibit uniform distributions, where rare samples primarily affect R@1 retrieval performance., leading to a larger improvement in the R@1 metric. The proposed MELT achieves better semantic denoising effects on the fashion domain dataset compared to the open-domain dataset.

\subsection{Ablation Study}\label{sec:ablation studies}
\noindent To assess the contribution of each component in the model, we conduct a detailed ablation study on the modules of the MELT model. The specific group design is as follows:
\textit{\textbf{G[A]: Ablation on  Rarity-Aware Token Refinement}}
\textbf{D\#1 wo\_R\_att}: We removed the attention block and replaced it with average pooling of image tokens to obtain the regions of interest for calculating the rarity score.
\textbf{D\#2 wo\_R\_thr}: We remove the threshold for rare sample classification and treat all samples as rare samples for modification.
\textbf{D\#3 wo\_R\_mod}: We remove the modification to the image tokens to investigate the contribution of the Rarity-Aware Token Refinement module.
\textit{\textbf{G[B]: Ablation on Diffusion-Based Similarity Denoising}}
\textbf{D\#4 wo\_D\_nes}: We remove Gaussian noise to investigate the contribution of noise addition to the denoiser training.
\textbf{D\#5 wo\_D\_$\mathcal{L}_{diff}$}: We remove the constraint loss $\mathcal{L}_{diff}$ from the diffusion module to investigate the contribution of the semantic denoiser.
\textbf{D\#6 wo\_D\_$\mathcal{L}_{KD}$}: We remove the knowledge distillation loss $\mathcal{L}_{KD}$ from the similarity matrix to investigate the contribution of the Diffusion-Based Similarity Denoising module.

\begin{table}[ht]
\caption{Ablation study on FashionIQ and CIRR datasets.}
\vspace{-8pt}
  \centering
  \tabcolsep=10pt
  \small
          \resizebox{\linewidth}{!}{
    \begin{tabular}{c|c|cc|cc}
    \hline
    \hline
    \multicolumn{1}{c|}{\multirow{2}{*}{\textbf{D\#}}} & \multirow{2}{*}{\textbf{Derivatives}} & \multicolumn{2}{c|}{\textbf{FIQ-Avg.}} & \multicolumn{2}{c}{\textbf{CIRR-Avg}}  \\
\cline{3-6}    \multicolumn{1}{c|}{} &       & \textbf{R@$10$} & \textbf{R@$50$} & \textbf{R@$k$} & \textbf{R$_{sub}$@$k$}\\
    \hline
    \multicolumn{6}{c}{\textit{\textbf{G[A]: Rarity-Aware Token Refinement}}} \\
    \hline
    1     & wo\_R\_att & 57.99 & 79.55 & 80.66 & 89.88 \\
    2 & wo\_R\_thr & 58.24 & 79.75 & 80.98 & 89.48  \\
    3 & wo\_R\_mod & 56.67 & 78.36 & 79.82 & 88.51  \\
    \hline
    \multicolumn{6}{c}{\textit{\textbf{G[B]: Diffusion-Based Similarity Denoising:}}} \\
    \hline
    4     & wo\_D\_nes & 58.04 & 79.79 & 80.74 & 89.67 \\
    5     & wo\_D\_$\mathcal{L}_{diff}$ & 57.50 & 79.47 & 80.54 & 89.66 \\
    6 & wo\_D\_$\mathcal{L}_{KD}$ & 58.20 & 79.61 & 80.78 & 89.83  \\

    \hline
    \hline
    \multicolumn{2}{c|}{\textbf{MELT (Ours)}} & \textbf{59.20} & \textbf{79.86} & \textbf{81.02} & \textbf{90.03}  \\
    \hline
    \hline
    \end{tabular}%
      }
      \vspace{-10pt}
  \label{tab:ablation}%
\end{table}%

As shown in Table~\ref{tab:ablation}, we obtain the following observations. Compared to the full MELT model: 1) The performance of \textbf{D\#(1)} and \textbf{D\#(2)} decreases, indicating that the attention block is crucial for accurately identifying the modification regions in the image guided by text and measuring the rarity of modification relationships. Additionally, properly judging the rarity of samples and making modifications significantly enhances the model's multimodal semantic understanding. The performance of \textbf{D\#(3)} shows a larger decline compared to \textbf{D\#(1)} and \textbf{D\#(2)}, highlighting the importance of optimizing rare samples, which reflects the contribution of the Rarity-Aware Token Refinement module.
2) The performance of \textbf{D\#(4)} shows a slight decrease, indicating that the addition of Gaussian noise effectively aids in the noise estimation from hard negative samples. The performance of \textbf{D\#(5)} declines more significantly, showing that the semantic denoiser with loss optimization can accurately capture semantic noise and exclude interference. The performance of \textbf{D\#(6)} also decreases, suggesting that the Diffusion-Based Similarity Denoising module effectively reduces the interference from high-frequency hard negative samples, significantly improving the model's robustness.
3) FashionIQ exhibits a larger drop than CIRR, indicating that the former contains more rare modification samples and hard negative samples. Overall, each component plays a positive role in the model's performance.
\section{Conclusion}
\noindent

In this work, we investigated two key limitations of existing CIR methods: frequency bias leading to ``Rare Sample Neglect'' and similarity scores being prone to interference from hard negative samples and noise. To address these limitations, we are confronted with two challenges: asymmetric rare semantic localization and robust similarity estimation under hard negative samples.To tackle these challenges, we proposed a novel CIR model, Modification frEquentation-rarity baLance neTwork , which allocates more attention to rare modification semantics in multimodal semantics and performs diffusion denoising on hard negative samples with high similarity, thereby facilitating multimodal composition and matching processes. Our MELT model achieved the optimal performance across all metrics on two CIR benchmark datasets.

\section{Acknowledgments}
This work was supported in part by the National Natural Science Foundation of China, No.:62276155, No.:62576195, and No.:62376137; in part by the China National University Student Innovation \& Entrepreneurship Development Program, No.:2025282 and No.:2025283.
\bibliographystyle{IEEEbib}
\bibliography{strings}

\end{document}